\documentclass[sigconf,screen]{acmart}

\AtBeginDocument{%
  \providecommand\BibTeX{{%
    \normalfont B\kern-0.5em{\scshape i\kern-0.25em b}\kern-0.8em\TeX}}}

\usepackage{algorithm}
\usepackage{algorithmic}
\usepackage{multicol}
\usepackage{multirow}

\renewcommand\footnotetextcopyrightpermission[1]{} 
\begin{document}

\title{Multi-clue Consistency Learning to Bridge Gaps Between General and Oriented Object in Semi-supervised Detection}

\author{Chenxu Wang}
\affiliation{%
  \institution{Nanjing University of Science and Technology}
  \city{Nanjing}
  \country{China}}
\email{facias914@gmail.com}

\author{Chunyan Xu}
\affiliation{%
  \institution{Nanjing University of Science and Technology}
  \city{Nanjing}
  \country{China}}

\author{Ziqi Gu}
\affiliation{%
  \institution{Nanjing University of Science and Technology}
  \city{Nanjing}
  \country{China}}

\author{Zhen Cui}
\affiliation{%
  \institution{Nanjing University of Science and Technology}
  \city{Nanjing}
  \country{China}}

\begin{abstract}
While existing semi-supervised object detection (SSOD) methods perform well in general scenes, they encounter challenges in handling oriented objects in aerial images. 
We experimentally find three gaps between general and oriented object detection in semi-supervised learning: 
1) Sampling inconsistency: the common center sampling is not suitable for oriented objects with larger aspect ratios when selecting positive labels from labeled data. 
2) Assignment inconsistency: balancing the precision and localization quality of oriented pseudo-boxes poses greater challenges which introduces more noise when selecting positive labels from unlabeled data. 
3) Confidence inconsistency: there exists more mismatch between the predicted classification and localization qualities when considering oriented objects, affecting the selection of pseudo-labels. 
Therefore, we propose a Multi-clue Consistency Learning (MCL) framework to bridge gaps between general and oriented objects in semi-supervised detection. 
Specifically, considering various shapes of rotated objects, the Gaussian Center Assignment is specially designed to select the pixel-level positive labels from labeled data. 
We then introduce the Scale-aware Label Assignment to select pixel-level pseudo-labels instead of unreliable pseudo-boxes, which is a divide-and-rule strategy suited for objects with various scales. 
The Consistent Confidence Soft Label is adopted to further boost the detector by maintaining the alignment of the predicted results. 
Comprehensive experiments on DOTA-v1.5 and DOTA-v1.0 benchmarks demonstrate that our proposed MCL can achieve state-of-the-art performance in the semi-supervised oriented object detection task. 
The code will be available at \href{https://github.com/facias914/sood-mcl}{https://github.com/facias914/sood-mcl}
\end{abstract}

\maketitle

\section{Introduction} \label{Introduction}

The fully-supervised object detection in aerial images~\cite{xia2018dota,han2021redet,xie2021oriented,li2023large} has achieved remarkable success in the past few years. 
However, labeling a large amount of accurate annotations is labour-consuming, especially for oriented dense objects. 
To reduce the annotation burden, the semi-supervised object detection (SSOD) attempts to leverage both labeled data and unlabeled data for boosting the detector performance.  
Existing advanced SSOD methods, which mainly adopt the self-training framework~\cite{liu2021unbiased,li2022pseco,liu2022unbiased,wang2023consistent} to employ these unlabeled data, have achieved significant performance in general scenes. 
However, they encounter various challenges in handling the oriented object detection task due to the tricky characteristics such as arbitrary rotation angles, large aspect ratio changes and dense arrangement. 
This prompts us to study the performance gaps between general and oriented object detection when performing semi-supervised learning. 

With comprehensive investigations in Section~\ref{Gaps Analysis}, we figure out that the performance of SOOD is significantly hindered by three gaps and inconsistency problems. 
Specifically, the sampling inconsistency represents that the positive labels selected from labeled data are inconsistent with the supervision information required by the oriented objects. 
It is caused by the sub-optimal center-based label assignment and the gap in aspect ratio distribution between general and oriented objects. 
That is to say, oriented objects usually have more extreme aspect ratio distributions than general ones, while the center-sampling strategy only focuses on the center region, which results in the discriminative information of large aspect ratio objects distributed at the edge being assigned as background, affecting the optimization of the detector. 

\begin{figure*}[ht] %
\vspace{-0.2cm}
	\footnotesize
	\centering
	\includegraphics[width=0.95\linewidth]{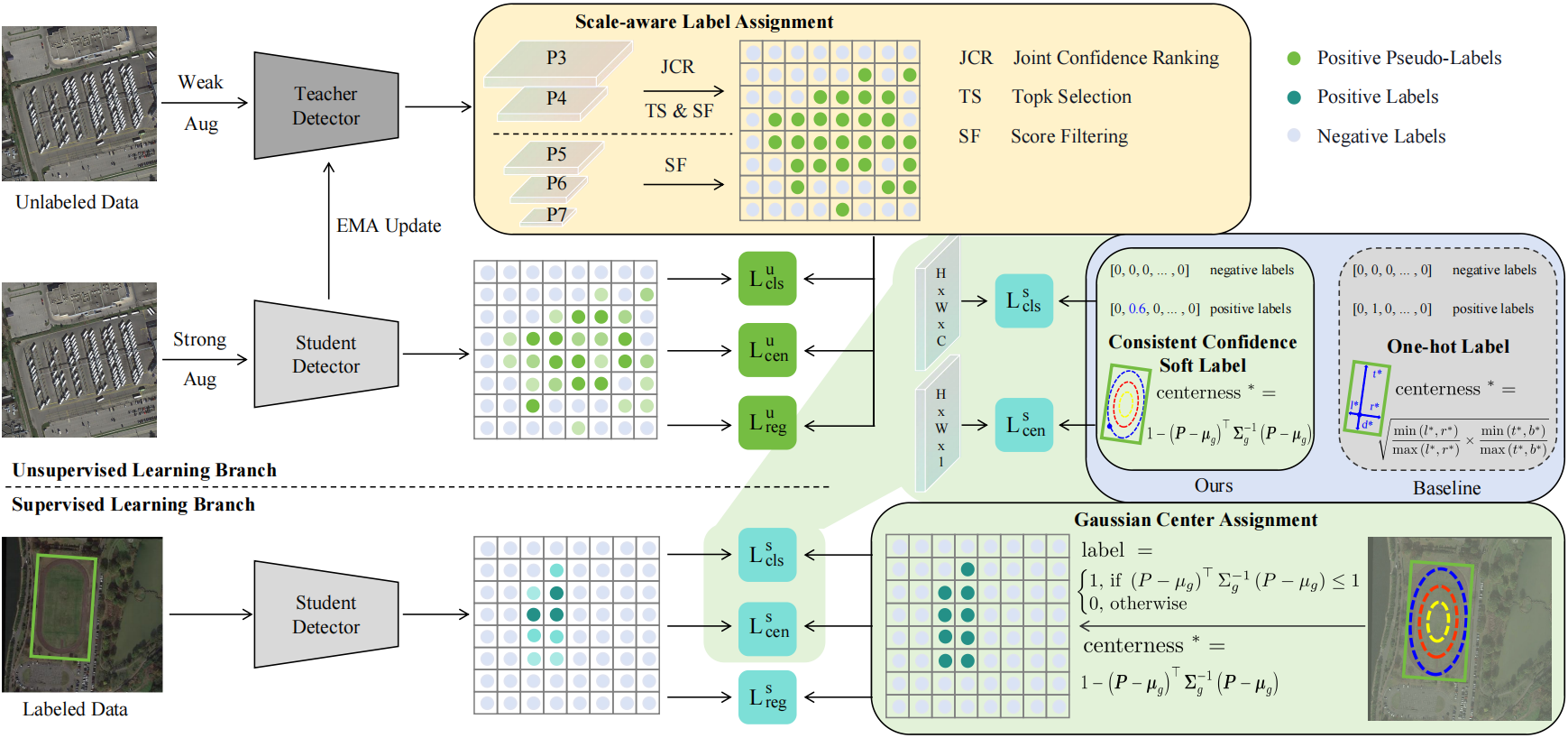} 
	\caption{The pipeline of our MCL. To address the sampling inconsistency issue, the Gaussian Center Assignment is introduced to select more accurate pixel-level positive labels from labeled data. The Scale-aware Label Assignment is proposed to select pixel-level pseudo-labels for objects with various scales. The Consistent Confidence Soft Label is adopted to mitigate the mismatch problem between classification and localization qualities through maintaining the alignment of the predicted results. 
}
	\vspace{-0.3cm}
	\label{pipline}
\end{figure*} 

In addition, the assignment inconsistency denotes that the predicted oriented pseudo-boxes from unlabeled data are difficult to guide the semi-supervised learning process, resulting in noisy label assignment.  
In general, most advanced SSOD methods~\cite{li2022pseco,wang2023consistent,zhang2023mind} follow the pseudo-boxes paradigm~\cite{sohn2020simple} that using the pseudo-boxes predicted by teacher model act as the supervised signals to optimize the student model. 
However, the introduction of angle parameters in rotated pseudo-boxes makes their localization quality less controllable than that of horizontal ones. 
Under the guidance of low-quality rotated pseudo-boxes, noisy labels are inevitably introduced during the label assignment process. 
Moreover, even with the aid of NMS (Non-Maximum Suppression) and threshold filtering, it is impossible to guarantee there are not redundant boxes which will bring additional noisy labels. 

On the other hand, the confidence inconsistency indicates that there exists the mismatch between the predicted classification and localization qualities, affecting the selection of pseudo-labels. 
Prior researches~\cite{sun2021makes,xu2021end,li2022pseco} have validated that the classification score is unable to measure the localization quality, and the lack of interaction between the two confidences causes an inconsistency in predictions, leading to sub-optimal pseudo-label selection results. 
Moreover, while numerous researches have studied the proxy localization quality of horizontal box, there is a lack of research for oriented boxes. 
Our experiments show that there remains a gap in the representation of localization quality between horizontal and oriented boxes, while sub-optimal proxy localization quality also affects the performance of oriented object detection. 

Based on the above observations, we propose a Multi-clue Consistency Learning (MCL) framework to boost the performance of the SOOD task through bridging multi-gaps between general and oriented object detection in semi-supervised learning process. 
Figure~\ref{pipline} illustrates the pipeline of MCL. 
Specifically, to address the sampling inconsistency problem, we introduce the Gaussian Center Assignment (GCA) strategy to select more accurate positive labels from the limited annotated data. 
Here each oriented object region can be represented as a $2D$ Gaussian Distribution~\cite{yang2022detecting} according to its corresponding ground truth box, which can be adopted to guide the pixel-level positive label selection.
In addition, the normalized distance to the center of the object (termed centerness) can be formulated as a $2D$ Gaussian distribution, which is used to better measure the localization quality of each pixel-level positive label. 
Compared to other label assignment strategies~\cite{ming2021dynamic,hou2022shape,xu2022rfla,xu2023dynamic} for fully-supervised detection, our designed GCA can not only extract more comprehensive target information from limited labeled data but also avoid the performance degradation caused by the sampling inconsistent supervision of large-ratio aerial objects. 

To mine accurate potential information from a large amount of unlabeled data, we would resist the assignment and confidence inconsistency. 
As for the assignment inconsistency, we propose the Scale-aware Label Assignment (SLA) method to select pixel-level pseudo-labels instead of unreliable pseudo-boxes, which adopts a divide-and-rule strategy to develop different pseudo-labels selection rules for objects with different scales. 
For small objects, a coarse-to-fine pseudo-label selection is used to filter high quality pseudo labels from dense features, which can preserve only a few candidates with both high classification and localization qualities. 
For large-scale objects with sparse features, we just use a score threshold filtering mechanism to maintain scale balance. 
Different from the previous dense pseudo-labels assignments~\cite{zhou2022dense,liu2023ambiguity} used in general scenes, our proposed SLA is more suitable for various scale objects in oriented object detection and then select more reliable pixel-level pseudo-labels from the unlabeled data. 

To address the confidence inconsistency problem, 
we employ the soft label~\cite{li2020generalized} that the value at the ground-truth category index indicates its corresponding localization quality, replacing the one-hot label to supervise the classification branch learning. 
However, there is a gap in the proxy representations of localization quality between horizontal and oriented boxes. 
Through experimental analysis in Section~\ref{Gaps Analysis}, we find that centerness~\cite{tian2019fcos} can effectively represent the localization quality of the oriented box and outperforms the predicted IoU (Intersection over Union) used on the horizontal box~\cite{li2020generalized,liu2023ambiguity}. 
Thereby we propose the Consistent Confidence Soft Label (CCSL) based on centerness to further promote the selection of pseudo-labels through mitigating the inconsistency problem. 
Prior to this, we have not seen any work on the localization quality representation of oriented box. 
In conclusion, our contributions are summarized as follows: 
\begin{itemize} 
    \item We look deep into the gaps between general and oriented object detection in semi-supervised learning, and then propose a multi-clue consistency learning framework to improve the performance of SOOD task.  
    \item To address three inconsistency problems, we specially design the Gaussian Center Assignment, the Scale-aware Label Assignment, and the Consistent Confidence Soft Label methods to mine more instructive supervised signals from the annotated data and unlabeled training data. 
    \item Comprehensive evaluations and comparisons demonstrate that our MCL achieves state-of-the-art performance on DOTA-v1.5 and DOTA-v1.0 benchmark~\cite{xia2018dota}. 
\end{itemize}

\section{RELATED WORK}

\textbf{Semi-supervised Object Detection.} Most of previous methods~\cite{liu2021unbiased,yang2021interactive,li2022rethinking,li2022semi,liu2023mixteacher,zhang2023mind,nie2023adapting} are inherited from the Mean Teacher paradigm~\cite{tarvainen2017mean}, where the teacher model is updated via Exponential Moving Average (EMA) of the student weights and produce pseudo labels over unlabeled data as ground truth for training the student model. 
Under this scheme, the quality and precision of pseudo labels play a substantial role. 
ISMT~\cite{yang2021interactive} maintains a memory bank to ensure the consistency of pseudo labels across various iteration stages. 
Unbiased Teacher~\cite{liu2021unbiased} applies Focal Loss~\cite{lin2017focal} to tackle the class-imbalance pseudo labels issues. 
Soft Teacher~\cite{xu2021end} employs a box jittering argumentation to select reliable pseudo labels. 
For misleading instances, Unbiased Teacher v2~\cite{liu2022unbiased} selects pseudo labels through utilizing uncertainty predictions.
Consistent Teacher~\cite{wang2023consistent} dynamically adjusting the threshold for pseudo labels filtering. 
Mix Teacher~\cite{liu2023mixteacher} enhances the quality of pseudo labels through mixed-scale prediction. 
Different from these, some methods~\cite{zhou2022dense,li2022dtg,liu2023ambiguity} directly use the dense predictions of the teacher model as pixel-level pseudo-labels for semi-supervised learning. 
However, since the usage scenarios of oriented object detection are more complex, there still exist the performance gaps when applying these SSOD methods in semi-supervised oriented object detection. 

\noindent \textbf{Semi-supervised Oriented Object Detection.} 
Oriented object detection requires predicting rotated bounding boxes by adding an angle parameter in the regression task, posing significant challenges for controlling the quality and accuracy of pseudo boxes. 
SOOD~\cite{hua2023sood} uses the absolute distance of the predicted angle between the teacher and student models to weight the regression loss. 
Additionally, it introduces the optimal transport cost~\cite{monge1781memoire} to evaluate the global similarity of layouts between the predictions of teacher and student. 
However, SOOD fails to recognize the root issue hindering semi-supervised learning in oriented object detection, which lies in how to eliminate inconsistency issues caused by the divergence of general and oriented object detection. 
In this work, we dive into the performance gaps between SSOD and SOOD methods, proposing a multi-clue consistency learning method for semi-supervised learning in oriented object detection. 

\noindent \textbf{Label Assignment in Oriented Object Detection.} Previous works have ~\cite{zhang2020bridging,ge2021ota,kim2020probabilistic} revealed that label assignment plays a crucial role in object detection. 
For oriented object detection, DAL~\cite{ming2021dynamic} introduce matching degree to select anchors with high prior and regression qualities. 
SASM~\cite{hou2022shape} set dynamic IoU thresholds decided by object shapes to filter prior anchors. 
In order to achieve scale-balance learning for tiny objects, RFLA~\cite{xu2022rfla}replaces the IoU-based or center sampling assignment with hierarchical label assignment built on gaussian respective field, while DCFL~\cite{xu2023dynamic} develops the dynamic priors instead of static priors.
Although these assignment methods are effective, they are limited to fully-supervised settings. 
With such over-designed label assignment strategies, the model is insufficient to learn perfect target information from sparse labeled data. 
In this work, we show that a simple changed sampling prior significantly improves the performance of SOOD task. 

\noindent \textbf{Representation of Localization Quality.} The representation of localization quality has attracted increasing attention in general object detection task. 
FCOS~\cite{tian2019fcos} utilize a separate branch to perform localization quality estimation in the form of centerness. 
After that, GFL~\cite{li2020generalized} points out that centerness is easier to get very small values in object edge by its definition, making it perform consistently worse than IoU. 
In SSOD task, several researches~\cite{liu2022unbiased, li2022pseco, liu2023ambiguity} also focus on accurately measuring the localization quality of pseudo labels. 
For instance, PseCo~\cite{li2022pseco} measures the localization quality of pseudo labels by estimating the consistency of predicted proposals~\cite{ren2015faster} for two-stage detectors. 
Unbiased Teacher v2~\cite{liu2022unbiased} reveals that the centerness is not reliable for reflecting whether a prediction is a positive sample due to lack of background supervision. 
Thereby it uses the predicted localization uncertainty of four box boundaries instead. 
ARSL\cite{liu2023ambiguity} also replaces the centerness branch with the IoU branch following the VFNet\cite{zhang2021varifocalnet} and GFL\cite{li2020generalized}. 
However, none of these works are designed for rotating bounding boxes in oriented object detection. 
In this work, we figure out that centerness is still the best choice as the proxy localization quality and we will verify this in Section~\ref{Gaps Analysis}. 

\section{METHODOLOGY} 
To study the gaps between the general and oriented object detection in the semi-supervised learning, we start from the classic SSOD framework~\cite{tarvainen2017mean}, as introduced in Section~\ref{DPLF}. 
We then experimentally analyze the three inconsistency issues (Section~\ref{Gaps Analysis}). 
To mitigate the inconsistency and bridge the gaps, we subsequently propose a Multi-clue Consistency Learning (MCL) framework for the SOOD task. 
It mainly comprises three modules: Gaussian Center Assignment (Section~\ref{GCA}), Scale-aware Label Assignment (Section~\ref{SLA}), and Consistent Confidence Soft Label (Section~\ref{CCSL}).

\subsection{The Basic SSOD Framework} \label{DPLF}

We introduce the classic Pseudo-labeling SSOD framework inherited from Mean Teacher~\cite{tarvainen2017mean} as our baseline, which adopts a rotated FCOS version~\cite{tian2019fcos} as the detector. 
In general, the SSOD framework comprises two branches: one dedicated to the supervised learning branch and the other to the unsupervised learning branch. 
In the burn-in stage, the student model undergoes pre-training with the limited annotated data, where the supervised loss $\mathcal{L}^{s}$ is minimized by combining classification, centerness, and regression.  
Simultaneously, the learned weights of the student model are mirrored onto the teacher model.
In the self-training stage, the unlabeled training data with weak data augmentation is fed into the teacher model to generate pseudo-labels. 
Subsequently, the student model is further optimized with these generated pseudo-labels (i.e., supervised by $\mathcal{L}^{u}$), while the input comprises unlabeled data augmented more aggressively. 
Meanwhile, the supervised learning with $\mathcal{L}^{s}$ is maintained on the student network. The overall loss function is thus formulated as $\mathcal{L} = \mathcal{L}^{s}+\beta \mathcal{L}^{u}$, 
where $\beta$ is a hyper-parameter regulating the impact of unsupervised learning.  
The network parameters of the student model are continually updated onto the teacher network through an exponential moving average (EMA) mechanism at each iteration.

\subsection{Inconsistency Analysis} \label{Gaps Analysis} 

\begin{figure}[t]
  \centering
  \setlength{\abovecaptionskip}{0.cm}
  \includegraphics[width=\linewidth]{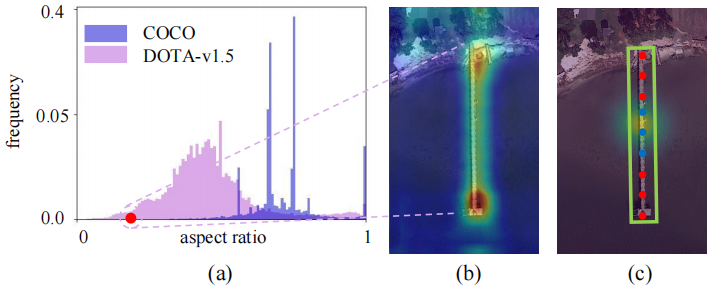}
  \caption{Analysis of the sampling inconsistency on the general COCO and aerial DOTA-v1.5 datasets. (a) Statistics of object aspect ratio distribution on both datasets. (b) The class activation mapping of an aerial object with large aspect ratio. (c) The positive label selection of an aerial object by the common center sampling strategy, where the red points and blue points represent negatives and positives, respectively. }
  \label{ratio}
  \vspace{-0.3cm}
\end{figure}

Since oriented objects in aerial images have some specific characteristics (e.g., the bird’s-eye view perspective, the complex environment, and the variant scales of objects), the standard SSOD framework would encounter various challenging problems when directly applied to the SOOD task. 
We attempt to analyze the  core components of the SSOD framework (including positive label assignment and pseudo-label selection), and then find some gaps/inconsistencies between general and oriented object detection in the semi-supervised learning. 
Specifically, we conduct a series of experimental analysis on two datasets representing general scenes and aerial perspectives, namely COCO~\cite{lin2014microsoft} and DOTA-v1.5~\cite{xia2018dota}, respectively. 
To ensure a fair comparison, we  use the FCOS as the object detector on both datasets.

\noindent \textbf{Sampling Inconsistency:}  
When employing these limited annotated data, the common center sampling strategy is adopted to select positive labels for general objects, but it is not suitable for oriented objects with larger aspect ratios.  
For an intuitive understanding, we statistically analyze the aspect ratio distributions of targets on the COCO and DOTA-v1.5 datasets, as shown in Figure~\ref{ratio}(a). 
More than 90\% of objects are with the aspect ratios exceeding 0.5 on the COCO dataset, thus the center sampling strategy can be adopted to select positive points of the general objects, thereby providing sufficient supervision signals. 
However, more than 70\% of objects exhibit aspect ratios less than 0.5 on the DOTA-v1.5 dataset, especially for some objects with extreme aspect ratios (seen in Figure~\ref{ratio}(b)) in the bird’s-eye view perspective. 
As seen in Figure~\ref{ratio}(c), only these blue points can be selected as pixel-level positive points with the common center sampling strategy, while these unselective red points would be seen as negative background information, leading to incorrect guidance for the model optimization. 
Therefore, the inconsistency between the shape-agnostic center sampling and objects with large aspect ratios would cause the performance degradation in the SOOD task. 

\begin{figure}[t]
  \centering
  \setlength{\abovecaptionskip}{0.cm}
  \includegraphics[width=\linewidth]{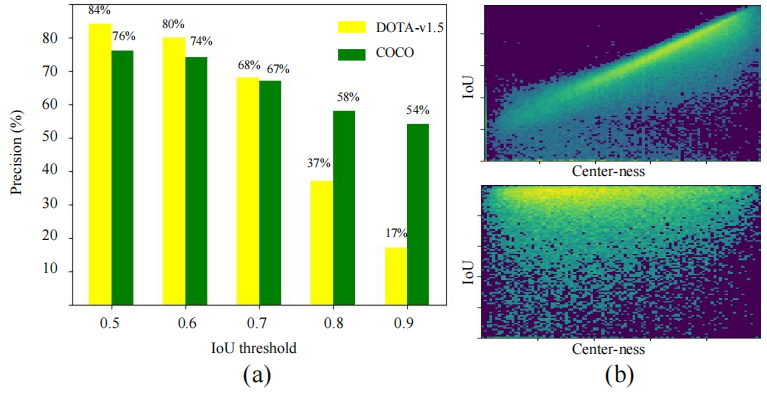}
  \caption{Investigation on the assignment and confidence inconsistency problems. (a) The pseudo-box precision of the DOTA-v1.5 dataset and COCO dataset under different IoU thresholds. (b) The top heat-map illustrates the consistency of centerness and IoU between ground truth boxes and their corresponding true positive boxes on the DOTA-v1.5 dataset, while the bottom one is on the COCO dataset.}
  \label{dota_coco}
  \vspace{-0.3cm}
\end{figure}

\noindent \textbf{Assignment Inconsistency.}  
Compared with the horizontal object detection in the general images, the localization of oriented objects in aerial images is more challenging due to their arbitrary rotation angles. 
To elucidate this, we conduct evaluations with the FCOS detector on the DOTA-v1.5 and COCO datasets, and then utilize the Intersection over Union (IoU) metric between predicted boxes and the corresponding ground-truth boxes to assess the localization quality of the pseudo-boxes. 
As shown in Figure~\ref{dota_coco}(a), we use different IoU thresholds to obtain the pseudo-box precision under the same recall rate on the DOTA-v1.5 and COCO datasets, respectively. 
With a lower IoU threshold, we can achieve a higher precision on the DOTA-v1.5 dataset compared to the COCO dataset, but the pseudo-boxes with the poor localization quality would affect the detection optimization.  
Under the guidance of low-quality rotated pseudo-boxes, noisy labels are inevitably introduced during the label assignment process. 
When improving the requirement of localization quality by increasing the IoU threshold (from 0.5 to 0.9) on both datasets, the detection precision would severely drop from 84\% to 17\% on the DOTA-v1.5 dataset, while it will be also reduced from 76\% to 54\% on the COCO dataset. 
A large number of redundant noisy boxes caused by reduced precision can also introduce the noisy labels. 
Hence, it's hard to achieve a trade-off between the precision and localization quality of rotated pseudo-boxes, and the assignment inconsistency problem should be addressed to guide the semi-supervised learning process.

\noindent \textbf{Confidence Inconsistency:} 
Prior research~\cite{li2020generalized,liu2023ambiguity} has validated that the confidence inconsistency problem is existing between the predicted classification and localization qualities. 
To enhance the correlation between them, most previous horizontal object detectors~\cite{li2020generalized,liu2023ambiguity} use the IoU-based soft label to supervise the classification branch. 
However, predicting a confidence IoU value is difficult for the oriented objects due to the increased uncertainty introduced by the rotation angle parameter. 
Meanwhile, we discover that centerness can well represent the localization quality of oriented bounding boxes.
As in Figure~\ref{dota_coco}(b), we statically analyze the IoU and centerness between all ground truth boxes and their corresponding true positive boxes on the DOTA-v1.5 dataset (the top subfigure) and the COCO dataset ((the bottom subfigure). Compared to the COCO dataset, the correlation between the IoU and centerness values is significantly higher on the DOTA-v1.5 dataset. 
Therefore, to address the confidence inconsistency problem, we utilize the predicted centerness value to measure the localization quality of oriented objects. 

\subsection{Gaussian Center Assignment} \label{GCA}
To deal with the sampling inconsistency problem, we attempt to select more suitable positive labels from the following aspects: 1) making full use of sparse annotated data; 2) taking the shape information of oriented objects into consideration. 
For the first condition, the optimal solution is to leverage these sparse annotations to assign more positive samples, hence some carefully designed label assignment strategies are excluded. 
Also considering the second condition, we propose a shape-aware method named Gaussian Center Assignment (GCA) to selectpixel-level positive labels for oriented object detection in semi-supervised learning. 

Concretely, We model the  object as a $2D$ Gaussian distribution that can well reflect the shape and direction of the aerial object. 
We denote a rotated bounding box as $B=\left(x_{c}, y_{c}, w, h, \theta \right)$, where $\theta$ represents the angle parameter; $(x_{c}, y_{c})$, $w$, and $h$ represent the center coordinates, width, and height of the oriented box, respectively. 
The coordinate of the center point $(x_{c}, y_{c})$ serves as the mean vector of $2D$ Gaussian distribution  $\boldsymbol{\mu}_{g}=(x_{c}, y_{c})^{\top}$, and the co-variance matrix $\boldsymbol{\Sigma}_{g}$ can be formulated as: 
\begin{equation}
    \boldsymbol{\Sigma}_{g}=\left[\begin{array}{cc}
    \cos \theta & -\sin \theta \\
    \sin \theta & \cos \theta
    \end{array}\right]\left[\begin{array}{cc}
    \frac{w^{2}}{4} & 0 \\
    0 & \frac{h^{2}}{4}
    \end{array}\right]\left[\begin{array}{cc}
    \cos \theta & \sin \theta \\
    -\sin \theta & \cos \theta
    \end{array}\right]. 
\end{equation} 
We then define a point coordinate as $\boldsymbol{P} = (x, y)^{\top}$ and determine whether it is a positive label based on the following formula: 
\begin{equation}
    \text { label }=\left\{\begin{array}{ll}
    1, & \text { if } (\boldsymbol{P} - \boldsymbol{\mu}_{g})^{\top} \boldsymbol{\Sigma}_{g}^{-1} (\boldsymbol{P} - \boldsymbol{\mu}_{g}) \leq 1 \\
    0, & \text { otherwise }.
    \end{array}\right.
\end{equation}
This label assignment ensures that positive labels cover not only the central region but also the edge region of the target, thereby avoiding the omission of discriminative features for aerial objects with large aspect ratios. 
Moreover, a large number of sub-optimal positive samples (pixels near the boundaries) also provide sufficient supervision information for the detector optimization. 
For the oriented object, the new expression of centerness is formulated as follows:
\begin{equation}
    \text { centerness }^{*}=1-(\boldsymbol{P} - \boldsymbol{\mu}_{g})^{\top} \boldsymbol{\Sigma}_{g}^{-1} (\boldsymbol{P} - \boldsymbol{\mu}_{g}). 
\end{equation}
According to the prior that pixels located around the center of the object are more representative than those near the object boundaries, we establish a new normalized distinction in the object region to represent the pixel-wise localization quality. 
The Gaussian Center Assignment (GCA) can select comprehensive supervision signals from the sparse annotated data to guide the learing, and further improve the performance for objects with large aspect ratios. 

\subsection{Scale-aware Label Assignment}  \label{SLA} 

\begin{figure}[t]
  \centering
  \setlength{\abovecaptionskip}{0.cm}
  \includegraphics[width=\linewidth]{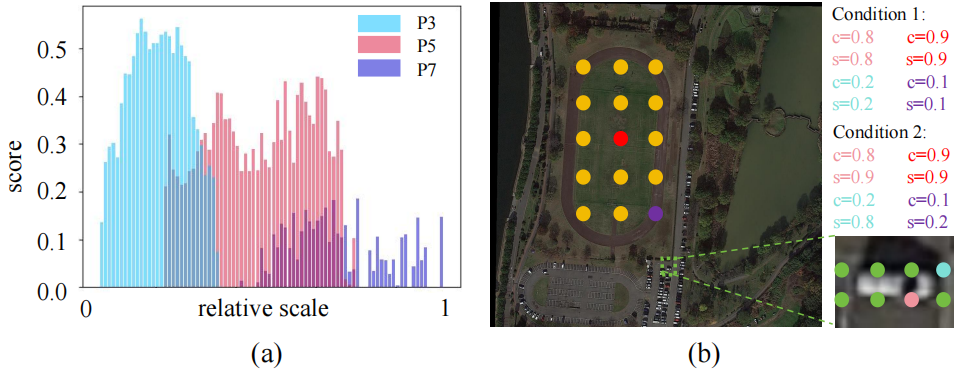}
  \caption{Analysis on difference feature maps and the object centerness. (a) Score imbalance between feature maps at different levels. (b) For centerness-based soft label, condition 1 introduce ambiguities while condition 2 is more suitable. }
  \label{score&scale}
  \vspace{-0.3cm}
\end{figure}

To tackle the assignment inconsistency problem, we conduct the pixel-level pseudo-label selection based on the feature level predictions of the teacher model rather than the post-processed pseudo-boxes. 
According to multi-level prediction with FPN~\cite{lin2017feature}, different sizes of objects can be detected on different levels of feature maps (i.e., $P_{3}, P_{4}, P_{5}, P_{6}, P_{7}$). 
However, as shown in Figure~\ref{score&scale}(a), there is a score imbalance problem between feature maps at different levels, where the higher level feature map tend to predict the lower scores. 
To balance the allocation of pseudo-labels for oriented targets with different scales, we propose Scale-aware Label Assignment (SLA) to improve the quality of pseudo-labels by employing different sampling rules for different level feature maps. 

For the low-level feature maps (i.e., $P_{3}, P_{4}$) with dense features, we design a coarse-to-fine rule to predict these small-scale objects with two following stages. 
In the coarse selection stage, the joint confidence (denoted as the multiplication of score and centerness) is used to rank all pixels and the top-k sorted pixels will be selected as candidate pseudo-labels. 
The joint confidence is expected to select samples with the high classification and localization qualities. 
As has been proven that the joint confidence is dominated by centerness due to its higher value than classification score~\cite{liu2022unbiased}, a high centerness value is unreliable to determine whether a pixel is positive since the centerness branch lacks supervision from negative instances. 
Therefore, we further filter these candidates through a score threshold to ensure the precision. 

For the high-level feature maps (i.e., $P_{5}, P_{6}, P_{7}$) with the lower predicted scores, we only use the score threshold to filter high-confidence pseudo-labels, thereby avoiding the unbalanced pseudo-labels allocation across various scales. Here the unsupervised loss consists of three parts: 
\begin{equation}
    \mathcal{L}_{u}=
    \frac{1}{N_{all}} \sum_{i}^{N_{all}} \mathcal{L}_{i}^{cls} + \frac{\alpha}{N_{pos}} \sum_{j}^{N_{pos}} w_{j} (\mathcal{L}_{j}^{cen}+\mathcal{L}_{j}^{reg})
\end{equation}
where $\alpha$ is a weighting parameter, $N_{all}$ represents the number of pixels across five level feature maps, $N_{pos}$ denotes the number of selected pseudo-labels. 
We apply the quality focal loss~\cite{li2020generalized}, the binary cross-entropy loss and the smooth-l1 Loss for guiding the classification, centerness and regression branches, respectively. 
Besides,  the localization weights $w$ are determined by the following formula: 
\begin{equation} 
    w_{j}=\left\{\begin{array}{ll}
    s_{j} \times c_{j} & j \in [P_{3}, P_{4}] \\
    s_{j} & j \in [P_{5}, P_{6}, P_{7}]
    \end{array}\right.
\end{equation} 
where $s$ and $c$ denote the score and centerness, respectively.
Since the pseudo-labels selection is based on model predictions, the existence of false positive labels is inevitable and the localization task is more sensitive to such noisy samples. 
Through employing the weight $w$, we reduce the contribution of low-confidence samples to the overall loss. 

\subsection{Consistent Confidence Soft Label} \label{CCSL} 
In this part, we focus on alleviating the confidence inconsistency issue. 
Based on the analysis in Section~\ref{Gaps Analysis}, we design a centerness-based soft label to supervise the classification branch. 
The soft label have to satisfy two conditions: 
(1) being positively correlated with centerness; 
(2) the score variance among all points within the target should not be too large, especially for small targets. 
Therefore,  we propose the  Consistent Confidence Soft Label (CCSL) strategy to align the classification and localization confidences. 

The core idea of CCSL is to replace the value of one-hot label at corresponding category index with a float value $y \in [0, 1]$, which satisfies the condition that the covariance calculated with centerness equal to 1. 
The soft label $y \in [0, 1]$ is given as follows: 
\begin{equation}
    \text { y }= [1-(\boldsymbol{P} - \boldsymbol{\mu}_{g})^{\top} \boldsymbol{\Sigma}_{g}^{-1} (\boldsymbol{P} - \boldsymbol{\mu}_{g})]^{\gamma}
\end{equation}
where $\gamma$ is a scale factor designed for the second condition. 
As shown in Figure~\ref{score&scale}(b), all points belonging to a small object have smaller stride, resulting in very similar point features. 
We hope that the score values of all points within the target are very close. 
Therefore, we weight them with the scale factor $\gamma$, which is formulated as $\sqrt[\beta]{(h \times w)/(H \times W)}$. $h$ and $w$ represent the height and width of the oriented box, $H$ and $W$ denote the height and width of the whole image, while $\beta$ controls the smoothing degree of $\gamma$. 
$\gamma$ can help prevent confusion during model training caused by excessive differences in point scores within the target. 
The proposed CCSL can bridge the gap between horizontal and oriented boxes in localization quality, and effectively mitigate the mismatch between classification and localization confidences, thereby bolstering the semi-supervised learning performance. 

\section{Experiment} \label{Experiments}

\subsection{Experimental Setup}

\textbf{Dataset:} The experiments are conducted on DOTA-v1.5~\cite{xia2018dota} and DOTA-v1.0~\cite{xia2018dota}. 
DOTA-v1.5 comprises 16 categories which contains 400$k$ annotated oriented instances. 
DOTA-v1.5-train, DOTA-v1.5-val, and DOTAv1.5-test contain 1411, 458, and 937 images, respectively. 
DOTA-v1.0 uses the same images as DOTA-v1.5 and the number of annotated instances is half the DOTA-v1.5's, with targets having a pixel area smaller than 10 being ignored. 
We include three evaluation protocols: 
1) DOTA1.5-Partial. Following SOOD method~\cite{hua2023sood}, we randomly sample10\%, 20\%, and 30\% images from DOTA-v1.5-train as labeled data and set the remaining images as unlabeled data. 
2) DOTA1.5-Full. We set DOTA-v1.5-train as labeled data, DOTA-v1.5-test as unlabeled data and perform the evaluation on the DOTA-v1.5-val. 
3) DOTA1.0-Full. The DOTA-v1.0-train and DOTA-v1.0-test are employed as labeled data and unlabeled data respectively, while the DOTA-v1.0-val is set as the evaluation dataset. 
For all evaluation protocols, the mAP~\cite{xia2018dota} is adopted as the evaluation metric. 

\noindent \textbf{Implementation Details:} We adopt the FCOS~\cite{tian2019fcos} as the detector and ResNet-50~\cite{he2016deep} pretrained on ImageNet~\cite{deng2009imagenet} as the backbone. 
Following the previous works~\cite{li2023large,xie2021oriented,han2021redet}, we crop the original images into patches of size 1024 $\times$ 1024, with an overlap region of 200 pixels between adjacent patches. 
To ensure a fair comparison, the MCL is trained on 2 RTX4090 GPUs with 3 images per GPU (2 labeled images and 1 unlabeled image). 
The SGD optimizer is applied with an initial learning rate of 0.0025, a momentum of 0.9, and a weight decay of 0.0001. 
For 30\% partial and full setting experiments, the MCL is trained for 180$k$ iterations, and the learning rate is divided by 10 at 120k and 160k. 
For 10\% partial and 20\% partial setting experiments, we train the MCL for 120$k$ iterations, while the learning rate is divided by 10 at 80k and 110k. 
We follow the same pipeline in~\cite{hua2023sood,zhou2022dense}, employing a burn-in strategy to initialize the parameters of the teacher model and applying the EMA strategy to update its parameters. 
The weak data augmentation strategy for the teacher model and strong data augmentation strategy for the student model are also consistent with them~\cite{hua2023sood,zhou2022dense}. 

\begin{table}[t]
\caption{Performance comparison with other state-of-the-art methods on the DOTA-v1.5-Partial dataset. ${\circ}$ and ${\dagger}$ denotes the base detector is Faster RCNN and FCOS respectively.}
\vspace{-0.3cm}
\begin{tabular}{clccc}
\toprule
    &    & \multicolumn{3}{c}{mAP(\%) $\uparrow$} \\
    \cmidrule(l){3-5}
\multirow{-2}{*}{Task} & \multirow{-2}{*}{Method} & 10\% & 20\%  & 30\% \\
\midrule 
\multirow{2}{*}{OD}  &Faster RCNN~\cite{girshick2015fast}  & 43.43 & 51.32 & 53.14 \\
   &FCOS~\cite{tian2019fcos}         & 42.78 & 50.11 & 54.79 \\
\midrule 
\multirow{6}{*}{SSOD} &Unbaised Teacher$^{\circ}$~\cite{liu2021unbiased}  & 44.51 & 52.80 & 53.33 \\
   & Soft Teacher$^{\circ}$~\cite{xu2021end}     & 48.46 & 54.89 & 57.83 \\
   & Dense Teacher$^{\dagger}$~\cite{zhou2022dense}   & 46.90 & 53.93 & 57.86 \\
   & PseCo$^{\circ}$~\cite{li2022pseco}   & 48.04 & 55.28 & 58.03 \\
   & DualPolish$^{\circ}$~\cite{zhang2023mind}   & 49.02 & 55.17 & 58.44 \\
   & ARSL$^{\dagger}$~\cite{liu2023ambiguity}   & 48.17 & 55.34 & 59.02 \\
\midrule 
\multirow{3}{*}{SOOD} & SOOD*$^{\dagger}$~\cite{hua2023sood}   & 48.63 & 55.58 & 59.23 \\
   & PST$^{\circ}$~\cite{wu2024pseudo} & 49.63 & 57.39 & 60.40 \\
   & \textbf{MCL$^{\dagger}$(Ours)} & \textbf{52.98} & \textbf{59.63} & \textbf{62.63} \\
\bottomrule 
\end{tabular}
\label{dota15-partial}
\vspace{-0.3cm}
\end{table}

\begin{table}[t]
\caption{Performance comparison on DOTA-v1.5-Full and DOTA-v1.0-Full, while numbers in font of the arrow indicate the result of supervised baseline} 
\vspace{-0.3cm}
\begin{tabular}{lcc}
\toprule
Method & DOTA-v1.5-Full & DOTA-v1.0-Full\\ 
\hline
Unbiased Teacher~\cite{liu2021unbiased}   & 66.12 $\xrightarrow{-1.27}$ 64.85 & -\\
Soft Teacher~\cite{xu2021end}       & 66.12 $\xrightarrow{+0.28}$ 66.40 & -\\
Dense Teacher~\cite{zhou2022dense}      & 65.46 $\xrightarrow{+0.92}$ 66.38 & 66.72 $\xrightarrow{+3.58}$ 
                                                         70.30\\
SOOD*~\cite{hua2023sood}              & 65.46 $\xrightarrow{+2.24}$ 67.70 & 66.72 $\xrightarrow{+4.09}$ 
                                                         70.81\\
\textbf{MCL(Ours)}      & $\textbf{65.46} \xrightarrow{\textbf{+3.62}} \textbf{69.08}$ & 
                     $\textbf{66.72} \xrightarrow{\textbf{+6.91}} \textbf{73.63} $\\
\bottomrule 
\end{tabular}
\label{dota15-full&dota1-full}
\vspace{-0.5cm}
\end{table}

\subsection{Comparison with State-of-the-Arts}
To validate the effectiveness of MCL, we compare it with previous methods under DOTA1.5-Partial, DOTA1.5-Full and DOTA1.0-Full evaluation protocols. 
For distinguishing with SOOD task, the SOOD~\cite{hua2023sood} method is termed as SOOD*. 
Moreover, some SSOD methods with their oriented version also participate in comparison. 

\noindent \textbf{DOTA-v1.5-Partial:} Experiment results under the DOTA-v1.5-Partial protocol are presented in Table~\ref{dota15-partial}. 
In SOOD task, SSOD methods typically underperform compared to SOOD methods, which exhibits the necessity of bridging the gap between horizontal boxes and oriented boxes in semi-supervised detection. 
SOOD*~\cite{hua2023sood} and PST~\cite{wu2024pseudo} are specifically designed for the characteristics of oriented boxes and remote sensing images, yet the performance gains are marginal. 
By bridging the gaps and addressing the concerns of inconsistency, our MCL achieves remarkable enhancements over the supervised baseline FCOS and outperforms all SOOD methods under all proportions. 
It scores 52.98, 59.63 and 62.63 mAP on 10\% 20\% 30\%  experiments, respectively, largely surpassing the state-of-the-arts method PST by 2 $\sim$ 3.5 mAP, validating the necessity of our analyses and methods. 

\noindent \textbf{DOTA-v1.5-Full and DOTA-v1.0-Full:} 
Table~\ref{dota15-full&dota1-full} gives the comparison results in terms of mAP on the DOTA-v1.5-Full and DOTA-v1.0-Full configurations. 
Our MCL demonstrates remarkable advancements over existing methods on the DOTA-v1.5-Full and DOTA-v1.0-Full experiment settings. 
Compared to the state-of-the-art SOOD* using the same detector, the MCL can achieve the notable improvement: 1.38\% on the the DOTA-v1.5-Full setting, and 2.82\% on the DOTA-v1.0-Full setting, exhibit its effectiveness. 

\begin{figure*}[ht] %
	\footnotesize
	\centering
	\includegraphics[width=0.98\linewidth]{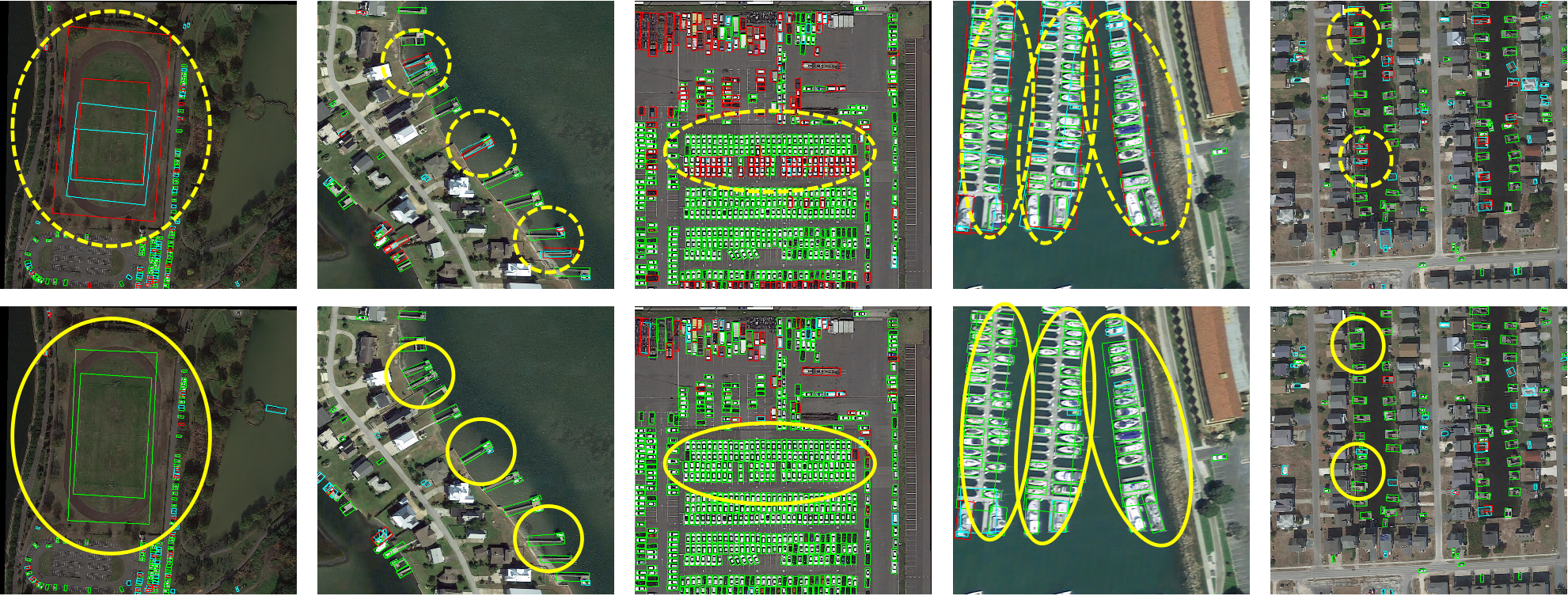} \vspace{-0.2cm} 
	\caption{Some visualization results from the DOTA-v1.5 dataset. The first and the last rows are the results of Dense Teacher and MCL respectively. True Positive, False Negative, and False Positive predictions are marked in green, red, and blue respectively. 
}
	\vspace{-0.2cm}
	\label{visual}
\end{figure*} 

\noindent \textbf{Visualization Analysis:} 
The visualization results under the 30\% setting of DOTA-v1.5-Partial are shown in Figure~\ref{visual}. 
Compared with the Dense Teacher~\cite{zhou2022dense}, MCL shows an enhancement in the detection of targets with large aspect ratios aided by the GCA module. 
Additionally, supported by the SLA module, the MCL demonstrates an improved capability in handling targets of various sizes in remote sensing scenarios. 
Moreover, the CCSL module assists the MCL in identifying high-quality pseudo-labels in densely populated target scenes, effectively reducing the incidence of False Negative predictions.
This collective integration of GCA, SLA, and CCSL within the MCL not only refines its detection precision but also establishes a robust approach for enhancing object detection in complex remote sensing images. 

\subsection{Ablation Studies}
Subsequently, we delve into the detailed analysis of each module. 
All the ablation experiments are performed on the 30\% setting of DOTA-v1.5-Partial without special instructions. 
We set the Dense Teacher~\cite{zhou2022dense} as the baseline. 
It also uses pixel-level pseudo labels by sorting all pixels according to classification scores and selects the top ratio(\%) ones. 
The original parameter setting sets ratio to 1\%, and we set ratio
to 3\% by default to maximize the performance. 
\begin{table}[t]
\caption{ Performance comparison with different modules. }
\vspace{-0.3cm}
\begin{tabular}{cccccccc}
\toprule
 &  &  &  & \multicolumn{3}{c}{mAP(\%) $\uparrow$} \\
    \cmidrule(l){5-7}
  & \multirow{-2}{*}{GCA} & \multirow{-2}{*}{SLA} & \multirow{-2}{*}{CCSL} & 10\% & 20\% & 30\% \\
\hline
baseline & $\times$ & $\times$ & $\times$ &49.71&55.64&59.65\\
\hline
  \multirow{4}{*}{MCL}& $\checkmark$ & $\times$ & $\times$ &51.12&56.74&60.62 \\
  & $\times$ & $\checkmark$ & $\times$ &51.84&57.16&61.32 \\
  & $\checkmark$ & $\checkmark$ & $\times$ &52.02&57.28&61.46 \\
  & $\checkmark$ & $\checkmark$ & $\checkmark$ &52.98&59.63&62.63 \\
\bottomrule
\end{tabular}
\label{modules}
\vspace{-0.5cm}
\end{table}

\noindent \textbf{The effect of each module:} 
We study the effectiveness of our proposed three modules (GCA, SLA, CCSL) under different experiment settings (i.e., 10\% 20\% 30\% ). 
The results are shown in Table~\ref{modules}. 
After comprehensively extracting supervision information from annotated data, GCA facilitates a performance enhancement for MCL. 
Furthermore, SLA enhances the quality of unsupervised information extracted from unlabeled data and also contributes to performance improvement. 
The CCSL can further improve the precision of pseudo-labels by mitigating the inconsistency between classification and localization confidence, thereby the participation of the CCSL leads to additional performance improvement for MCL. 

\noindent \textbf{All Sampling $\textbf{vs}$ GCA:} 
To evaluate the effective of the GCA in SOOD task, we compare GCA with the center sampling strategy used in FCOS~\cite{tian2019fcos} and the all sampling strategy that labels all pixels insides ground-truth boxes as positives and the remaining pixels as negatives. 
The detailed comparable results are shown in Table~\ref{ablation-gca-1}. 
Although the aim is to fully utilize labeled data to extract more positive supervision information, all sampling strategy still shows performance degradation compared to center sampling. 
We speculate that this is due to all sampling introducing excessive background at the target's edges. 
In contrast, GCA can provide sufficient positive supervision information while accounting for the target's aspect ratio without introducing excessive noise, thereby improving the performance.

\begin{table}[t]
\footnotesize
\begin{minipage}[c]{0.22\textwidth}
    \caption{Performance comparison between GCA and All Sampling strategy.} \vspace{-0.2cm}
    \label{ablation-gca-1}
    \scalebox{1.25}{
    \begin{tabular}{lccc}
        \toprule        
        Sampling strategies & mAP \\
        \hline
        Center Sampling & 59.65  \\
         All Sampling & 58.49  \\
         \textbf{GCA} & \textbf{60.62}  \\
        \bottomrule 
    \end{tabular}
    }
\end{minipage}
\hspace{0.03\linewidth}
\begin{minipage}[c]{0.22\textwidth}
\centering 
    \caption{Performance comparison between Mean Teacher and Dense Teacher.} \vspace{-0.2cm}
    \label{ablation-ica-2}
    \scalebox{1.25}{
    \begin{tabular}{lccc}
        \toprule        
        Method & mAP \\
        \hline
        FCOS(base) & 54.79  \\
        \hline
        Mean Teacher & 26.02 (-28.77)  \\
        Dense Teacher & 59.65 (+4.86)  \\
        \bottomrule 
    \end{tabular}
    }
\end{minipage}
\vspace{-0.3cm}
\end{table}

\begin{table}[t]
\caption{Ablation studies on SLA.}
\label{ablation-ica-1}
\vspace{-0.3cm}
\begin{tabular}{ccccccc}
\toprule
      & ratio & thr & topk & Recall & Precision & mAP \\
\hline
\multirow{2}{*}{score} & 1\% & - & -& 40.08  & 30.76  & 57.86 \\
                       & 3\% & - & -& 69.44  & 17.76  & 59.65 \\
\hline
score $\times$ center & 3\% & - & -& 70.38  & 18.01 & 60.45 \\
\hline
\multirow{6}{*}{SLA}& \multirow{3}{*}{-} & 0.01 & \multirow{3}{*}{1500} & 90.35  & 12.74 & 59.97 \\
    &  &  0.02 &  & 88.76  & 19.50     & 60.61 \\
    &  &  0.03 &  & 87.15  & 24.03     & 60.59 \\
\cmidrule(l){2-7}
   & \multirow{3}{*}{-} &  0.01    & \multirow{3}{*}{2000} & 93.72  & 12.39     & 60.34 \\
   &  & 0.02 &   & 91.89  & 19.13     & 61.32  \\
   &  & 0.03 &   & 90.04  & 23.69     & 60.89 \\
\bottomrule
\end{tabular}
\vspace{-0.5cm}
\end{table}

\noindent \textbf{The analysis of SLA:} 
To verify the impact of SLA hyper-parameters and demonstrate the advantages of SLA over other selection strategies, we conduct the experiments with the pixel-level recall and precision as the metrics. 
In detail, we denote the selected points from the ground truth boxes as ground truth points. 
The selected pseudo-labels that coincide with ground truth points are denoted as true positives and the remaining ones are false positives. 
The computation forms of pixel-level recall and precision are consistent with the previous work~\cite{lin2014microsoft}.  
The experimentation is shown in Table~\ref{ablation-ica-1}. 
It can be observed that the performance of semi-supervised learning improves 1.79\% mAP when recall increases from 40.08\% to 69.44\%, demonstrating that maintaining high recall of pseudo-labels is crucial for ensuring the performance of SOOD task. 
In addition, replacing score selection used in Dense Teacher~\cite{zhou2022dense} with joint confidence selection can slightly enhances both recall and precision, thereby improving the performance from 59.65 mAP to 60.45 mAP. 
However, it is still not the optimal solution. 
By considering the imbalance between scores of objects of different scales and the disparity between scores and centerness values, SLA significantly enhances recall while maintaining satisfactory precision, thereby markedly improving SOOD performance.

\noindent \textbf{The analysis of CCSL:} 
To validate our analysis in Section~\ref{Gaps Analysis} that predicted centerness is a more suitable proxy localization quality of oriented box than predicted IoU, we perform the comparative experiment between them. 
As shown in Table~\ref{ablation-ccsl-1}, compared to predicting the uncontrollable IoU, predicting the centerness with considering the scale information improves the SOOD performance. 
In addition, the scale factor $\gamma$ is also a critical component. 
The central and edge features of extremely small objects are very similar, leading to significant differences in the predicted centerness values confuse the algorithm's training. 
By combining the scale factor $\gamma$, the variance of the target internal center distribution can be dynamically adjusted according to the target scale. 
Therefore, compared to vanilla centerness, centerness with the scale factor can better enhance the detection performance.

\begin{table}[t]
\caption{Ablation studies on CCSL.}
\vspace{-0.3cm}
\begin{tabular}{lccc}
\toprule        
Localization Quality &  $\beta$  &  mAP \\
\hline
IoU &  -  & 60.98  \\
\hline
\multirow{4}{*}{Centerness} & 0 &  60.87  \\
     & 0.1 &  61.38  \\
    & 0.2 &  62.63  \\
     & 0.25 &  62.26  \\
\bottomrule 
\end{tabular}
\label{ablation-ccsl-1}
\vspace{-0.2cm}
\end{table}

\begin{table}[t]
\caption{The impacts of GCA on objects with large aspect ratios. }
\vspace{-0.3cm}
\begin{tabular}{cccccccc}
\hline
  & \multicolumn{2}{c}{HB} & \multicolumn{2}{c}{BD} & \multicolumn{2}{c}{GTF} & \multirow{2}{*}{mAP} \\
  & Recall  & mAP & Recall & mAP & Recall & mAP & \\
\hline
FCOS & 81.7 & 71.8 & 60.8 & 40.1 & 72.6 & 55.2 & 65.5 \\
+GCA  & 84.6 & 74.1 & 66.5 & 45.7 & 83.0 & 62.5 & 67.2 \\
\hline
\end{tabular}
\label{ablation-gca-3}
\vspace{-0.5cm}
\end{table}

\subsection{Inconsistency Mitigation}
\noindent \textbf{Sampling Inconsistency:} We conduct additional experiments to demonstrate the effectiveness of GCA in sampling inconsistency mitigation. 
The metric is the detection performance for objects with large aspect ratios. 
The models are trained on the DOTA-v1.5-train and evaluated on the DOTA-v1.5-val. 
We extract the three categories with the highest aspect ratio from the 16 categories in DOTA-v1.5, namely HB (harbor), BD (bridge), and GTF (ground-track-field). 
The experimental results are shown in Table~\ref{ablation-gca-3}. 
With the support of GCA, there is a notable enhancement in both recall and mAP for these objects with the large aspect ratios, which verifies that GCA can effectively alleviating the sampling inconsistency, enhancing the performance of large aspect ratio object detection. 

\noindent \textbf{Assignment Inconsistency:} 
To verify the necessity of using pixel-level pseudo-labels to eliminate assignment inconsistency as discussed in Section~\ref{Gaps Analysis}, 
we compare the performance of the pseudo-boxes-based method Mean Teacher(we follow the Consistent Teacher~\cite{wang2023consistent} that call the vanilla pseudo-boxes framework Mean Teacher) and the pixel-level pseudo-labels-based method Dense Teacher~\cite{zhou2022dense} on FCOS~\cite{tian2019fcos}. 
As shown in Table~\ref{ablation-ica-2}, Mean Teacher exhibits a significant performance drop due to the fact that dense object detectors adopt a binary label assignment strategy, making them highly sensitive to the overall quality of the oriented pseudo-boxes which is difficult to guarantee. 
In contrast, pixel-level pseudo-labels greatly alleviate the assignment inconsistency introduced during label assignment, thereby improving the performance. 

\begin{figure}[t]
  \centering
  \setlength{\abovecaptionskip}{0.cm}
  \includegraphics[width=\linewidth]{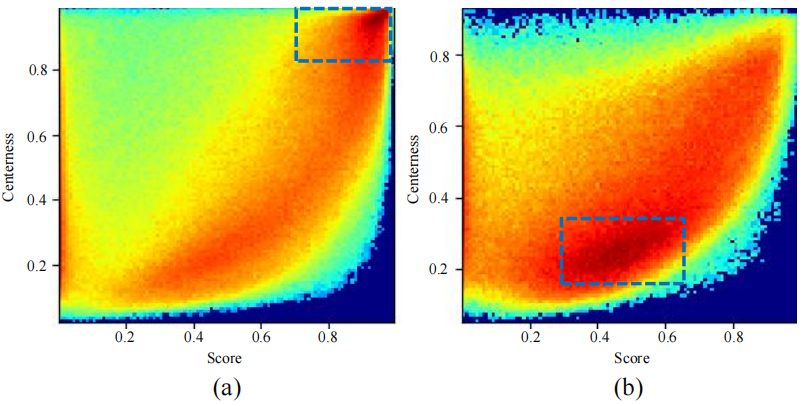}
  \caption{Illustrated analysis of CCSL. (a) The heatmap of the scores and centerness of positive samples trained based on CCSL. (b) The heatmap of the scores and centerness of positive samples trained based on one-hot label. }
  \label{ccsl-2}
\end{figure}

\noindent \textbf{Confidence Inconsistency:} 
To assess the effectiveness of CCSL in eliminate confidence inconsistency, we train two FCOS detectors on DOTA-v1.5-train, one of which is supervised with one-hot label as usual and the other with CCSL. 
We visualize the score-centerness heatmaps of all ground truth points on the DOAT-v1.5-val set, which is shown in Figure~\ref{ccsl-2}. 
As highlighted in the blue squares, with the incorporation of CCSL, there is an improvement in the positive correlation between classification scores and centerness that represents the localization quality, which demonstrates that the CCSL is effective in eliminating the inconsistency between classification and localization qualities. 

\section{Conclusion}

In this work, we propose a Multi-clue Consistency Learning (MCL) framework to boost the performance of semi-supervised oriented object detection. 
We attempt to bridge the gaps between general and oriented object detection in semi-supervised learning through investigating three inconsistency issues.
To mitigate the sampling inconsistency, we design the Gaussian Center Assignment to consider various shapes of aerial objects by employing the limited annotation data.  
The proposed Scale-aware Label Assignment and Consistent Confidence Soft Label strategies to mine more accurate supervised information from a large amount of unlabeled data, which can further promote the semi-supervised learning. 
Extensive experimental results on the public DOTA-v1.0 and DOTA-v1.0 datasets clearly demonstrate our proposed MCL is effective to bridge the gaps between the general and oriented object detection in the semi-supervised learning. 

\bibliographystyle{ACM-Reference-Format}
\bibliography{sample-base}

\end{document}